\documentclass[letterpaper]{article} 
\usepackage{aaai2027}  
\usepackage[hyphens]{url}  
\usepackage{graphicx} 
\usepackage{natbib}  
\usepackage{caption} 
\usepackage{algorithm}

\usepackage{graphicx}
\usepackage{booktabs}
\usepackage{amsmath}
\usepackage{amssymb}
\usepackage{mathtools}
\usepackage{multirow}
\usepackage{caption}
\usepackage{algpseudocode}
\usepackage{pifont}
\usepackage{tcolorbox}
\usepackage{enumitem}
\usepackage{tabularx}
\usepackage{array}
\usepackage{url}
\newcommand{\tit}[1]{\smallbreak\noindent\textbf{#1.}}

\usepackage{newfloat}
\usepackage{listings}
\DeclareCaptionStyle{ruled}{labelfont=normalfont,labelsep=colon,strut=off} 
\floatstyle{ruled}
\newfloat{listing}{tb}{lst}{}
\floatname{listing}{Listing}

\usepackage{booktabs}

\newcommand{\ours}{SceneExpander}

\title{SceneExpander: Text-Guided 3D Scene Expansion via Free-Form View Insertion}
\author{
    Written by AAAI Press Staff\textsuperscript{\rm 1}\thanks{With help from the AAAI Publications Committee.}\\
    AAAI Style Contributions by Peter Patel Schneider,
    Sunil Issar,\\
    J. Scott Penberthy,
    George Ferguson,
    Hans Guesgen,
    Francisco Cruz\equalcontrib\corresponding,
    Marc Pujol-Gonzalez\equalcontrib\corresponding
}
\affiliations{
    \textsuperscript{\rm 1}Association for the Advancement of Artificial Intelligence\\

    1101 Pennsylvania Ave, NW Suite 300\\
    Washington, DC 20004 USA\\
    proceedings-questions@aaai.org
}

\title{SceneExpander: Text-Guided 3D Scene Expansion via Free-Form View Insertion}
\author {
    Zijian He\textsuperscript{\rm 1},
    Renjie Liu\textsuperscript{\rm 1},
    Yihao Wang\textsuperscript{\rm 1},
    Weizhi Zhong\textsuperscript{\rm 1},\\
    Huan Yang\textsuperscript{\rm 2},
    Kun Gai\textsuperscript{\rm 2},
    Guangrun Wang\textsuperscript{\rm 1},
    Guanbin Li\textsuperscript{\rm 1}\corresponding
}
\affiliations {
    \textsuperscript{\rm 1}Sun Yat-sen University,
    \textsuperscript{\rm 2}KlingAI Research, Kuaishou Technology
}

\begin{document}

\twocolumn[{%
\renewcommand\twocolumn[1][]{#1}%
\maketitle

\begin{center}
    \captionsetup{
        type=figure,
        width=0.9\textwidth
    }
    \vspace{-42pt}
    \includegraphics[width=0.9\textwidth]{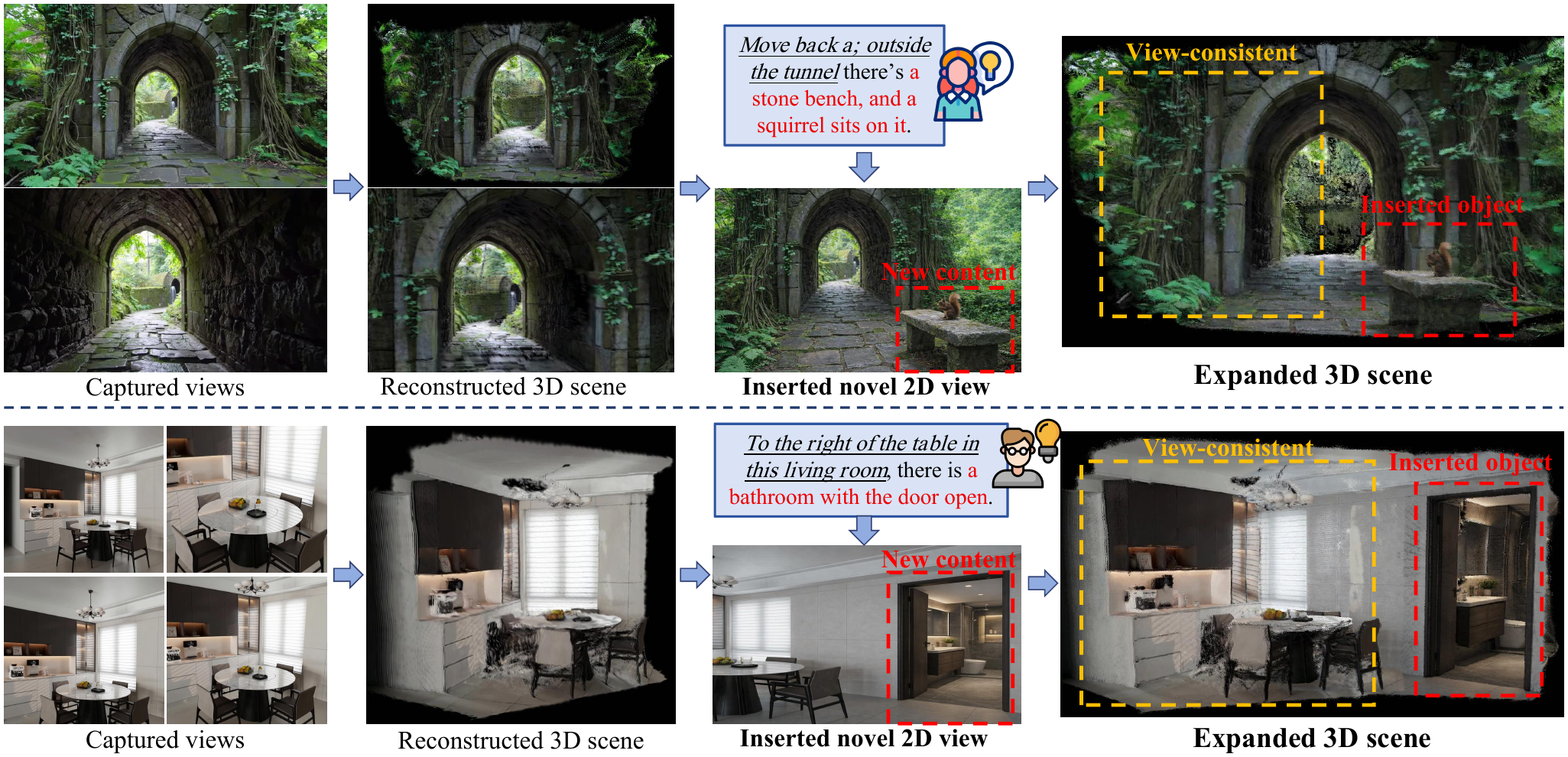}
    \vspace{-5pt}
    \captionof{figure}{
        \textbf{Two examples of controllable text-guided 3D scene expansion
        via free-form view insertion.}
        Starting from a 3D reconstruction of captured views, the user provides
        a text instruction describing a desired extension, which is materialized
        as an inserted novel view and integrated to expand the scene.
        Yellow marks the view-consistent captured area, and red highlights
        newly introduced content or objects. 
    }
    \label{fig:teaser}
\end{center}
}]

\begin{abstract}
World building with 3D scene representations is increasingly important for content creation, simulation, and interactive experiences, yet real workflows are inherently iterative: creators repeatedly extend existing scenes under user control. Motivated by this gap, we study \textbf{text-guided 3D scene expansion via free-form view insertion}. Starting from a real scene captured by multi-view images, a user specifies a text expansion intent, which a generative model materializes as an inserted view extending the scene coverage. Unlike simple object editing or style transfer within a fixed scene, the inserted view may be 3D-misaligned with the original reconstruction, introducing geometric shifts, hallucinated content, or view-dependent artifacts that disrupt global multi-view consistency. To address this challenge, we propose \ours, which applies test-time adaptation to a parametric feed-forward 3D reconstruction model with two complementary distillation signals: anchor distillation stabilizes the captured scene using geometric cues from the captured views, while inserted-view self-distillation retains insertion-supported predictions to accommodate the misaligned view. Experiments on ETH scenes and online data demonstrate improved expansion behavior and reconstruction quality under misalignment. Project page: \url{https://scnuhealthy.github.io/SceneExpander}.
\end{abstract}

\section{Introduction}
World models and 3D scene representations are increasingly central to \emph{world building}---the creation, completion, and extension of virtual environments for content production, simulation, and interactive experiences~\cite{schneider2025worldexplorer,huang2025voyager,kerbl20233d}. In practice, however, realistic world building rarely happens as a one-shot, prompt-to-world generation. Instead, creators typically construct scenes iteratively: starting from partial assets or captures, they progressively extend the environment, add new regions, and refine details under user control. A natural interface for this workflow is \textbf{text-guided 3D scene expansion}: a user describes how the existing scene should grow, and the system expands the reconstruction accordingly. Directly editing 3D content from text remains challenging, while image generative models provide an accessible way to materialize the text instruction as a free-form novel view~\cite{zhuang2024tip,haque2023instruct,wu2024gaussctrl,he2025vton,zhuang2023dreameditor,zheng2025splatpainter}. Motivated by this practical need, we introduce a user-centric workflow for controllable scene expansion (Fig.~\ref{fig:teaser}). Starting from a real scene captured by a set of multi-view images, a user provides a text instruction describing a desired extension, such as revealing an unseen region or continuing structures beyond the captured boundary. An image generative model materializes this instruction as an additional inserted view conditioned on the available scene context. The system then inserts this view into the reconstruction, enabling the scene to be expanded iteratively into a larger, editable 3D world.

\begin{figure*}[t]
    \centering
    \includegraphics[
        width=.9\textwidth
    ]{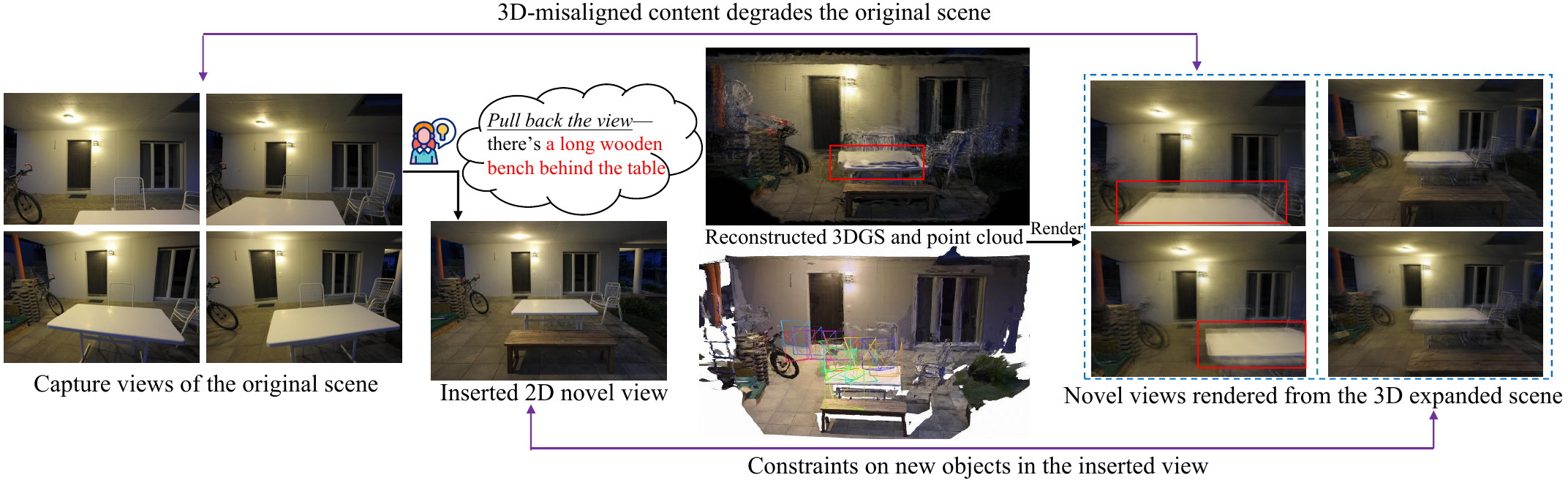}
    \caption{
        \textbf{Challenges in text-guided 3D scene expansion via free-form view insertion.}
        (1) A 3D-misaligned inserted view can corrupt reconstruction quality
        in the captured region.
        (2) Newly introduced content must remain consistent under novel
        viewpoints.
        Red boxes highlight ghosting artifacts caused by conflicting 3D
        constraints from the inserted view.
    }
    \vspace{-12pt}
    \label{fig:motivation}
\end{figure*}

However, unlike simple object edits or style transfer~\cite{zhuang2024tip,haque2023instruct,wu2024gaussctrl,he2025vton,zheng2025splatpainter}, inserting a view that introduces a new region inevitably creates spatial inconsistencies with the originally captured views. In this paper, we study an under-explored problem, \textbf{text-guided 3D scene expansion via free-form view insertion}: given a captured multi-view set of a real scene and an additional inserted view that may be \emph{3D-misaligned} with the captured geometry, our goal is to reconstruct an \emph{expanded} 3D scene that (i) preserves reconstruction quality and faithfulness to the original multi-view evidence in the captured region, while (ii) incorporates the inserted view as an expansion constraint to extend the scene beyond the observed boundary. As shown in Fig.~\ref{fig:motivation}, the core challenge is that the inserted view, although visually plausible, may exhibit geometric shifts or hallucinated content that cannot be explained by a single coherent 3D structure. If we enforce strict multi-view consistency, these conflicts can propagate through optimization, leading to unstable geometry and artifacts around the newly introduced content.

This capability, robustly integrating a text-instantiated but potentially 3D-misaligned inserted view to expand an existing reconstruction, remains largely unaddressed. Most feed-forward reconstruction methods~\cite{wang2025vggt,peng2025omnivggt,liu2025worldmirror} assume that all inputs originate from a single consistent scene; once a misaligned inserted view is added, they either suppress the new content or degrade the original reconstruction under conflicting multi-view constraints. Camera-guided video generation~\cite{dai2025fantasyworld,huang2025voyager} can synthesize coherent sequences, but it is primarily a forward pipeline and lacks a mechanism to expand a pre-captured scene by inserting new constraints. Another common response is to pursue strictly 3D-consistent view generation, yet enforcing such consistency is often costly, requiring substantial multi-view or 3D supervision and careful data curation, and still cannot guarantee geometric compatibility for arbitrary prompts and viewpoints. We therefore shift the focus from perfectly 3D-consistent text-conditioned generation to robust 3D integration: \emph{how can we expand a reconstructed 3D scene when the inserted view is not geometrically consistent with the existing geometry?}

To answer this question, we propose SceneExpander, a framework for expanding 3D scenes with text-guided, misaligned inserted views. Our key insight is that this setting is well suited to test-time adaptation: the captured views provide reliable instance-specific constraints, while the inserted view carries the user's expansion intent but is 3D-noisy,
making fixed feed-forward inference and hard supervision brittle. Test-time adaptation enables minimal, scene-specific updates from a pre-trained reconstructor to accommodate the new constraint without retraining or requiring 3D-consistent generation. Concretely, SceneExpander builds upon a feed-forward reconstruction model (WorldMirror~\cite{liu2025worldmirror}) and performs robust test-time tuning using two complementary distillation signals. Anchor distillation uses the camera parameters, depth maps, and normal maps predicted from the original multi-view set to stabilize the captured scene, and we further apply data augmentation to improve robustness on the original views. In parallel, inserted-view self-distillation adapts latent geometry and appearance to explain the inserted view when present while retaining observation-supported predictions. Together, these objectives yield an expanded reconstruction that remains faithful to the captured scene yet flexibly incorporates misaligned new evidence, enabling controllable scene expansion without requiring perfectly 3D-consistent generation.


Our contributions are three-fold:
(1) We introduce the task of text-guided 3D scene expansion via free-form view insertion, formulating a user-centric workflow that bridges text-driven view generation, inserted-view integration, and editable 3D world building.
(2) We propose SceneExpander, a test-time tuning framework that adapts a parametric feed-forward reconstruction model to text-guided, 3D-misaligned inserted views via anchor distillation and inserted-view self-distillation, enabling robust expansion while preserving the original captured scene.
(3) We demonstrate the effectiveness of SceneExpander on ETH scenes and online data, showing improved expansion behavior and reconstruction quality under misalignment.


\section{Related Work}

\subsection{3D Reconstruction}

Classical image-based reconstruction pipelines typically follow Structure-from-Motion (SfM) and Multi-View Stereo (MVS), with systems such as COLMAP widely used to estimate camera poses and dense geometry from image collections~\cite{schonberger2016structure}. Neural scene representations have recently advanced both reconstruction quality and novel-view rendering. NeRF~\cite{mildenhall2021nerf} optimizes a continuous radiance field from posed images and achieves high-quality view synthesis, while 3D Gaussian Splatting (3DGS)~\cite{kerbl20233d} provides an explicit yet optimizable representation with efficient training and real-time rendering. These optimization-based methods are highly effective when the input views are mutually consistent, but can become brittle when additional constraints contradict the original geometry.

In parallel, \emph{feed-forward} reconstruction models have rapidly evolved toward large, general-purpose 3D predictors. 
VGGT~\cite{wang2025vggt} demonstrates a transformer that directly infers multiple key 3D attributes (e.g., cameras, pointmaps, depths, tracks) from one to many views within seconds. Building on this, OmniVGGT~\cite{peng2025omnivggt} further enables omni-modality geometric prediction by injecting auxiliary geometric cues (e.g., depth and camera parameters) via lightweight adapters, thereby supporting flexible modality combinations at inference. TTT3R~\cite{chen2025ttt3r} focuses on dynamic scene reconstruction and improves long-sequence performance through test-time updates that modulate the recurrent memory using alignment confidence, leading to stronger length generalization. WorldMirror~\cite{liu2025worldmirror} likewise targets unified geometric prediction, taking optional priors (e.g., intrinsics, depth) as input and producing multiple 3D outputs (e.g., points, depths, normals, 3D Gaussians) within a single forward pass.
While these feed-forward models improve efficiency and robustness, they typically assume that all inputs originate from a single coherent scene. When an additionally inserted AI-generated view is \emph{misaligned} with the captured observations, naively conditioning on it can suppress the new content or destabilize the reconstruction, motivating our robust scene-expansion setting in SceneExpander.

\subsection{World Generation and Editing}

A growing body of work aims to generate explorable worlds by leveraging strong generative priors. WorldExplorer~\cite{schneider2025worldexplorer} synthesizes navigable scenes via iterative, trajectory-based generation and fuses the generated views into a 3DGS representation. However, WorldExplorer largely operates by interpolating viewpoints within the generated environment and does not explicitly support expanding beyond the originally observed region. 

Camera-guided world and video generation systems such as HunyuanWorld-Voyager~\cite{huang2025voyager} and FantasyWorld~\cite{dai2025fantasyworld} generate RGB-D along user-defined camera paths to facilitate reconstruction and exploration, while interactive world models such as Yume-1.5~\cite{mao2025yume} emphasize long-horizon, text-controlled world evolution. These approaches explicitly couple video generation and geometric prediction within unified pipelines, aiming to improve geometric consistency during generation. However, this tight coupling can make post-hoc editing difficult, as it often requires  modifying the entire generated video rather than editing a single inserted view.

Scene-generation methods synthesize multi-view content and lift it to 3D, such as Text2Room~\cite{hollein2023text2room} and text-driven NeRF scene generation~\cite{zhang2024text2nerf}. Meanwhile, multi-view diffusion research improves cross-view consistency during novel-view generation, e.g., training-free interpolated denoising in ViewFusion~\cite{yang2024viewfusion} or 3D-prior-warped diffusion in MVGenMaster~\cite{cao2025mvgenmaster}. These methods primarily reduce inconsistency during generation. In contrast, we address a more challenging setting in which a user-provided generated view may still be misaligned, and we seek to integrate it robustly into an existing reconstruction.

3D scene editing methods modify existing scenes using instruction-guided image edits and representation re-optimization, including NeRF-based editing~\cite{haque2023instruct} and 3DGS-based variants such as Instruct-GS2GS, GaussianEditor, VcEdit, EditSplat, and video-prior-based editing~\cite{igs2gs,chen2024gaussianeditor,wang2024view,lee2025editsplat,chen2025fast}. 
These methods mainly edit content within observed regions, whereas we target scene expansion via view insertion, where the inserted view may contradict the original multi-view geometry and must be integrated while preserving captured evidence.

\begin{figure*}[t]
  \centering
  \includegraphics[width=\linewidth]{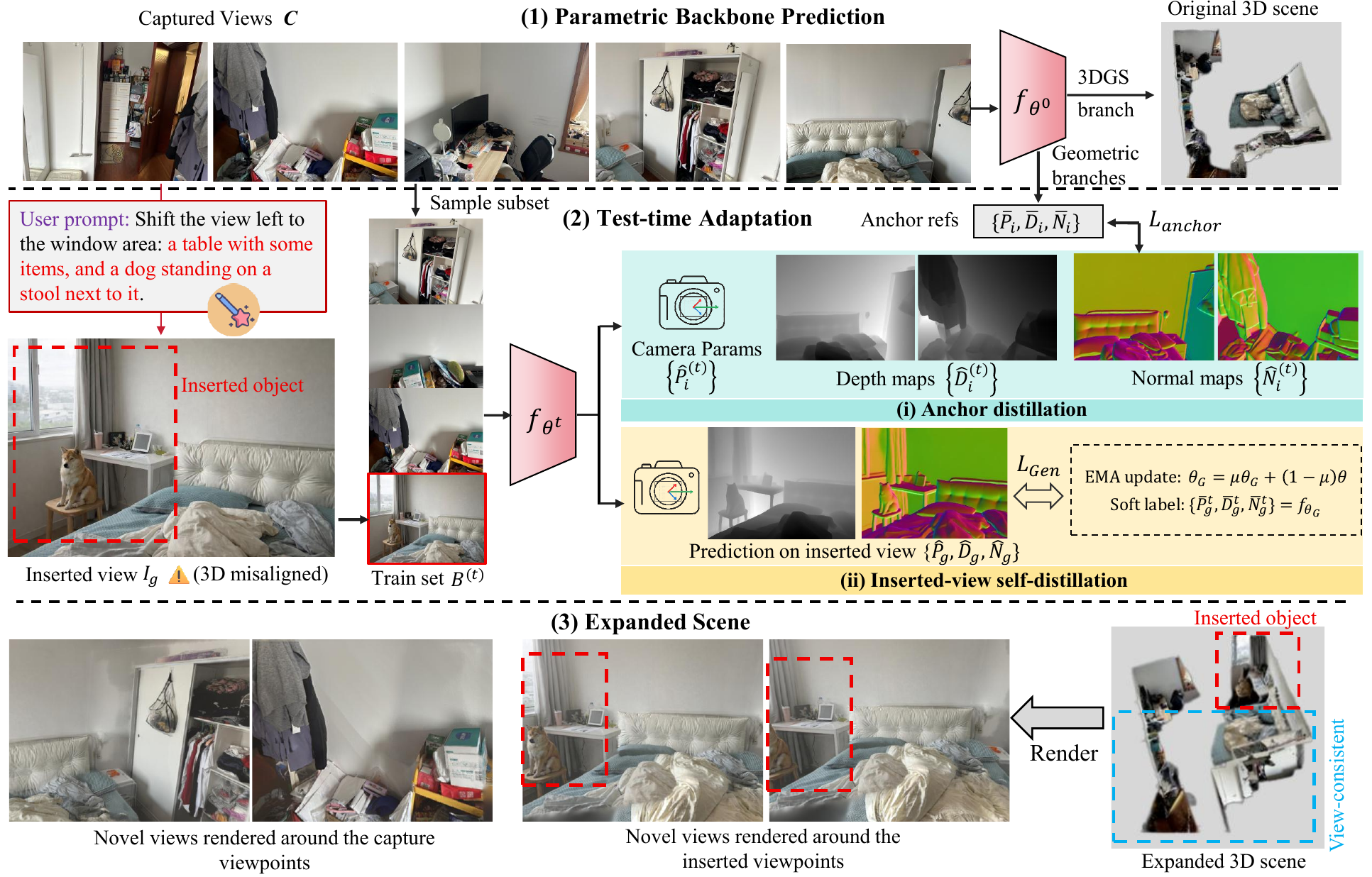}
  \vspace{-12pt}
  \caption{\textbf{Overview of SceneExpander.}
    (1) A user provides a text expansion instruction, which an image generative model materializes as a free-form inserted view $I_g$; a feed-forward reconstructor $f_\theta$ predicts an initial 3D scene from captured views.
    (2) To accommodate the text-guided inserted view that may be 3D-misaligned, we adapt $f_\theta$ at test time using two distillation losses: anchor distillation on captured views and inserted-view self-distillation on $I_g$.
    (3) The adapted model produces an expanded 3D scene consistent with the captured region while incorporating the inserted content.}

    \vspace{-10pt}
  \label{fig:method}
\end{figure*}

\section{Method}
\label{sec:method}

\subsection{Problem Setup}
\label{sec:problem_setup}

We study text-guided 3D scene expansion via free-form view insertion.
Starting from a real scene represented by $n$ captured multi-view images
$\mathcal{C}=\{I_i\}_{i=1}^{n}$, a user provides a text expansion instruction
$y$ describing how the scene should be extended, e.g., by revealing an
unseen region, moving toward a new viewpoint, or introducing new content.
An image generative model materializes this intent as a free-form inserted
view $I_g$, using the current scene context as reference.

Unlike conventional multi-view reconstruction, the inserted view is not
required to be strictly 3D-consistent with the captured observations.
It may contain pose shifts, hallucinated structures, or view-dependent
artifacts that cannot be fully explained by the original scene geometry.
Our goal is therefore to adapt a pre-trained multi-view reconstruction model
so that $I_g$ serves as a soft expansion constraint: the reconstructed scene
should incorporate the user-requested extension while preserving stable and
faithful predictions in the originally captured region.

In this formulation, text provides an intuitive high-level interface for
controlling scene expansion, while the inserted view supplies the visual
evidence through which the reconstruction is extended.

\subsection{SceneExpander}
\label{sec:tta_overview}
Directly treating the text-guided inserted view $I_g$ as a supervised target is brittle: hallucinated or shifted structures in the generated image can dominate the gradients and corrupt the reconstructed geometry. 
SceneExpander instead uses the pre-insertion model as a conservative prior and adapts it with two complementary distillation constraints:
(i) \textbf{anchor distillation} on captured views, and
(ii) \textbf{inserted-view self-distillation} with iteratively refined soft labels on $I_g$. The overview is shown in Fig.~\ref{fig:method}.

\subsubsection{Teacher--student setup.}
\label{sec:teacher_student}
Let $\theta_0$ denote the model parameters before inserting $I_g$. 
We instantiate three networks:
\begin{itemize}
    \item \textbf{Anchor teacher} $\theta_A \leftarrow \theta_0$ (frozen), which provides a fixed reference on captured views;
    \item \textbf{Generation teacher} $\theta_G^{(0)} \leftarrow \theta_0$, updated as an exponential moving average (EMA) of the student and providing soft labels on the generated inserted view;
    \item \textbf{Student} $\theta^{(0)} \leftarrow \theta_0$, which we adapt at test time.
\end{itemize}
At adaptation step $t$, the EMA generation teacher is updated as
\begin{equation}
\theta_G^{(t)} \leftarrow \mu\,\theta_G^{(t-1)} + (1-\mu)\,\theta^{(t)},
\label{eq:ema}
\end{equation}
where $\mu \in [0,1)$ is the momentum coefficient.
Thus the teacher predictions on $I_g$ form a sequence of \emph{soft labels} that are gradually refined as the student improves.

\subsubsection{Subset sampling and data augmentation.}
\label{sec:sampling}
\begin{figure}[t]
  \centering
  \includegraphics[width=\linewidth]{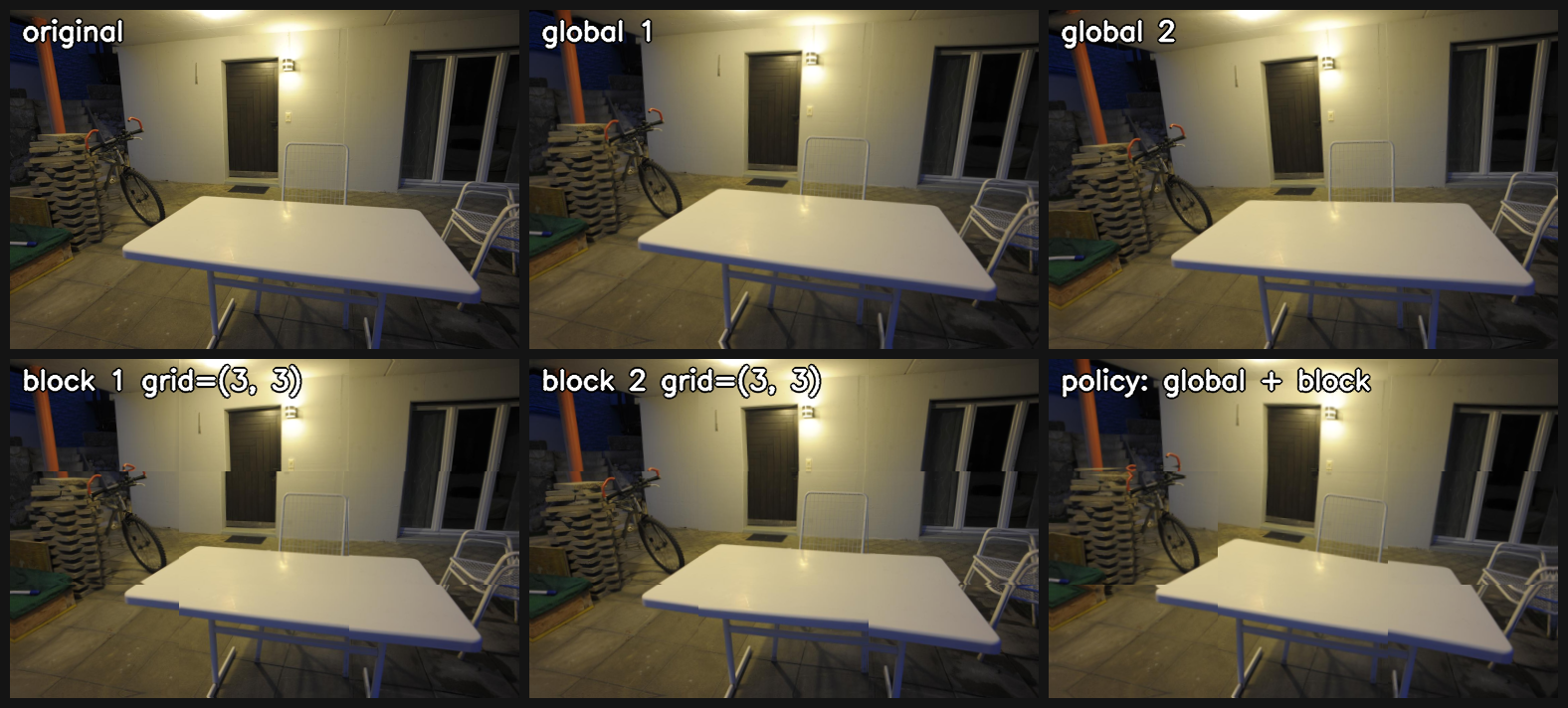}
  \caption{\textbf{Geometry-perturbation augmentation.}
We show four augmentation modes: identity (original), global affine warp, blockwise piecewise-affine warp (grid $3{\times}3$), and their combination (global+block), designed to mimic pose drift and local geometric inconsistencies.}
  \label{fig:geometry_aug}
  \vspace{-12pt}
\end{figure}

To stabilize test-time adaptation under 3D-misaligned insertion, as shown in Fig. \ref{fig:geometry_aug}, we construct each mini-batch by \emph{subset sampling} the captured views and applying a \emph{Geometry-Perturbation Augmentation}. This strategy exposes the model to diverse camera configurations while simulating spatial inconsistencies commonly introduced by generative inserted views. 
At each adaptation step $t$, we sample a subset of captured views from $\mathcal{C}$ with size in $\{n\!-\!1,\,n\!-\!2,\,n\!-\!3\}$, denoted by indices $\mathcal{V}^{(t)}$. We then optionally insert the generated view $I_g$ into the batch with probability $p$:
\begin{equation}
\mathcal{B}^{(t)} =
\begin{cases}
\{Aug(I_i)\}_{i\in\mathcal{V}^{(t)}} \cup \{I_g\}, & u < p,\\[2pt]
\{Aug(I_i)\}_{i\in\mathcal{V}^{(t)}}, & \text{otherwise},
\end{cases}
\quad u \sim \mathcal{U}(0,1),
\label{eq:batch}
\end{equation}
where $p{=}0.5$ by default.
Varying $\mathcal{V}^{(t)}$ forces the model to reconcile $I_g$ with different contextual captured views, discouraging overfitting to a single configuration and improving cross-view robustness.

Our augmentation \(Aug(\cdot)\) includes four variants: (i) identity, using the original image without modification; (ii) global-only affine perturbation; (iii) blockwise-only piecewise-affine perturbation; and (iv) global+block, applying a global warp followed by a blockwise warp. The global transform simulates overall pose drift, while the blockwise transform introduces weak local geometric perturbations by warping image blocks independently. We blend warped blocks with a smooth feathering mask to avoid seams. Together with subset sampling, Geometry-Perturbation Augmentation improves adaptation stability and reduces sensitivity to misalignment.

\subsubsection{Anchor distillation on captured views.}
\label{sec:anchor}
A central difficulty of view insertion is that the inserted image $I_g$ can be geometrically inconsistent: directly adapting the model to explain $I_g$ may introduce gradients that distort the already-correct reconstruction of the captured region. 
To prevent such catastrophic drift, we treat the pre-insertion model as a conservative prior and explicitly \emph{anchor} the reconstruction on the captured views. 
Concretely, we freeze an anchor teacher $\theta_A$ initialized from $\theta_0$ and precompute its per-view geometric predictions on the captured set:
\begin{equation}
(\bar{\mathbf{P}}_i,\bar{D}_i,\bar{N}_i) = f_{\theta_A}(\mathcal{C})\big|_{i},\qquad i=1,\ldots,n.
\label{eq:anchor_ref}
\end{equation}
At step $t$, given student predictions
\[
(\widehat{\mathbf{P}}_i^{(t)},\widehat{D}_i^{(t)},\widehat{N}_i^{(t)}) = f_{\theta^{(t)}}(\mathcal{B}^{(t)})\big|_{i},
\]
the anchor loss is
\begin{equation}
\begin{aligned}
\mathcal{L}_{\text{anchor}}^{(t)}
={}&
\sum_{i\in\mathcal{V}^{(t)}}
\Big(
d_{\text{cam}}(
\widehat{\mathbf{P}}_i^{(t)},
\bar{\mathbf{P}}_i)
\\
&\quad
+\alpha_D\, d_D(
\widehat{D}_i^{(t)},
\bar{D}_i)
+\alpha_N\, d_N(
\widehat{N}_i^{(t)},
\bar{N}_i)
\Big).
\end{aligned}
\label{eq:anchor}
\end{equation}

where $d_{\text{cam}}$ is a camera distance, $d_D$ is an $\ell_1$ (or scale-invariant) depth distance, and $d_N$ is a cosine normal distance.
This term enforces that adding $I_g$ does not change what the model believes about the captured scene.

\subsubsection{Inserted-view self-distillation with iteratively refined soft labels.}
\label{sec:gen}
The inserted view $I_g$ is synthesized and 3D-misaligned, so directly using its pixels as supervision is brittle. 
Instead, we integrate $I_g$ through self-distillation: an EMA teacher $\theta_G$ provides a soft pseudo-geometry target on $I_g$, and this target is progressively refined over adaptation steps. This makes $I_g$ act as a weak expansion constraint while avoiding overfitting to its inconsistencies.
At step $t$, if $I_g\in\mathcal{B}^{(t)}$, the student prediction on $I_g$ is
\begin{equation}
(\widehat{\mathbf{P}}_g^{(t)},\widehat{D}_g^{(t)},\widehat{N}_g^{(t)}) = f_{\theta^{(t)}}(\mathcal{B}^{(t)})\big|_{g}.
\label{eq:student_gen}
\end{equation}
The EMA teacher produces the corresponding soft label:
\begin{equation}
(\bar{\mathbf{P}}_g^{(t)},\bar{D}_g^{(t)},\bar{N}_g^{(t)})
=
f_{\theta_G^{(t-1)}}(\mathcal{B}^{(t)})\big|_{g}.
\label{eq:aug_avg}
\end{equation}
Since $\theta_G$ is updated as an EMA of the student (Eq.~\eqref{eq:ema}), the soft label evolves over time, gradually incorporating improved predictions from the adapted model.

We distill the student toward the soft label via
\begin{equation}
\begin{aligned}
\mathcal{L}_{\text{gen}}^{(t)}
={}&
d_{\text{cam}}\!\left(
\widehat{\mathbf{P}}_g^{(t)},
\bar{\mathbf{P}}_g^{(t)}
\right)
\\
&\quad
+\beta_D\, d_D\!\left(
\widehat{D}_g^{(t)},
\bar{D}_g^{(t)}
\right)
+\beta_N\, d_N\!\left(
\widehat{N}_g^{(t)},
\bar{N}_g^{(t)}
\right).
\end{aligned}
\label{eq:gen}
\end{equation}

and assign a small weight to $\mathcal{L}_{\text{gen}}^{(t)}$ so that $I_g$ serves as a soft constraint rather than hard supervision, enabling stable integration under misalignment.

\subsubsection{Stochastic restoration.}
\label{sec:restore}
To mitigate long-horizon drift under continuous adaptation, we additionally apply stochastic restoration every $K$ steps:
\begin{equation}
\theta^{(t)} \leftarrow \textsc{Restore}(\theta^{(t)},\theta_0;r),
\label{eq:restore}
\end{equation}
where $r$ controls the restoration rate.
This periodically pulls a random subset of parameters back toward the pre-insertion weights and acts as a safeguard against error accumulation on misaligned inputs.

\subsubsection{Full objective.}
\label{sec:full}
At each step $t$, we update the student by minimizing
\begin{equation}
\min_{\theta^{(t)}}\quad
\lambda_A\,\mathcal{L}_{\text{anchor}}^{(t)}
+
\lambda_G\,\mathcal{L}_{\text{gen}}^{(t)}
+
\lambda_{\text{reg}}\,\mathcal{L}_{\text{reg}}(\theta^{(t)}),
\label{eq:full}
\end{equation}
where $\lambda_A$ is set relatively large to preserve the captured region and $\lambda_G$ is small to avoid overfitting to the misaligned inserted view. The regularizer $\mathcal{L}_{\text{reg}}$ penalizes parameter drift during test-time adaptation, improving stability and reducing artifacts.
After $T$ adaptation steps, we obtain the adapted parameters $\theta^\star$.
Algorithm~\ref{alg:cotta_sceneexpander_compact} in Appendix summarizes the procedure.

\section{Experiments}
\label{sec:experiments}

\subsection{Experimental Setup}
\label{sec:setup}

\tit{Datasets}
\label{sec:datasets}
We evaluate on the ETH dataset~\cite{schops2017multi} and the online collection of WorldMirror~\cite{liu2025worldmirror}, covering outdoor and indoor scenes with wide-baseline multi-view observations. For each scene, we generate one instruction-guided novel view extending beyond the captured boundary (Fig.~\ref{fig:teaser}). We manually select a candidate that best follows the instruction; the selected view may still be 3D-misaligned with the captured scene. Results are reported on the merged ETH test set and online collection.

\begin{figure*}[t]
  \centering
  \includegraphics[width=.9\linewidth]{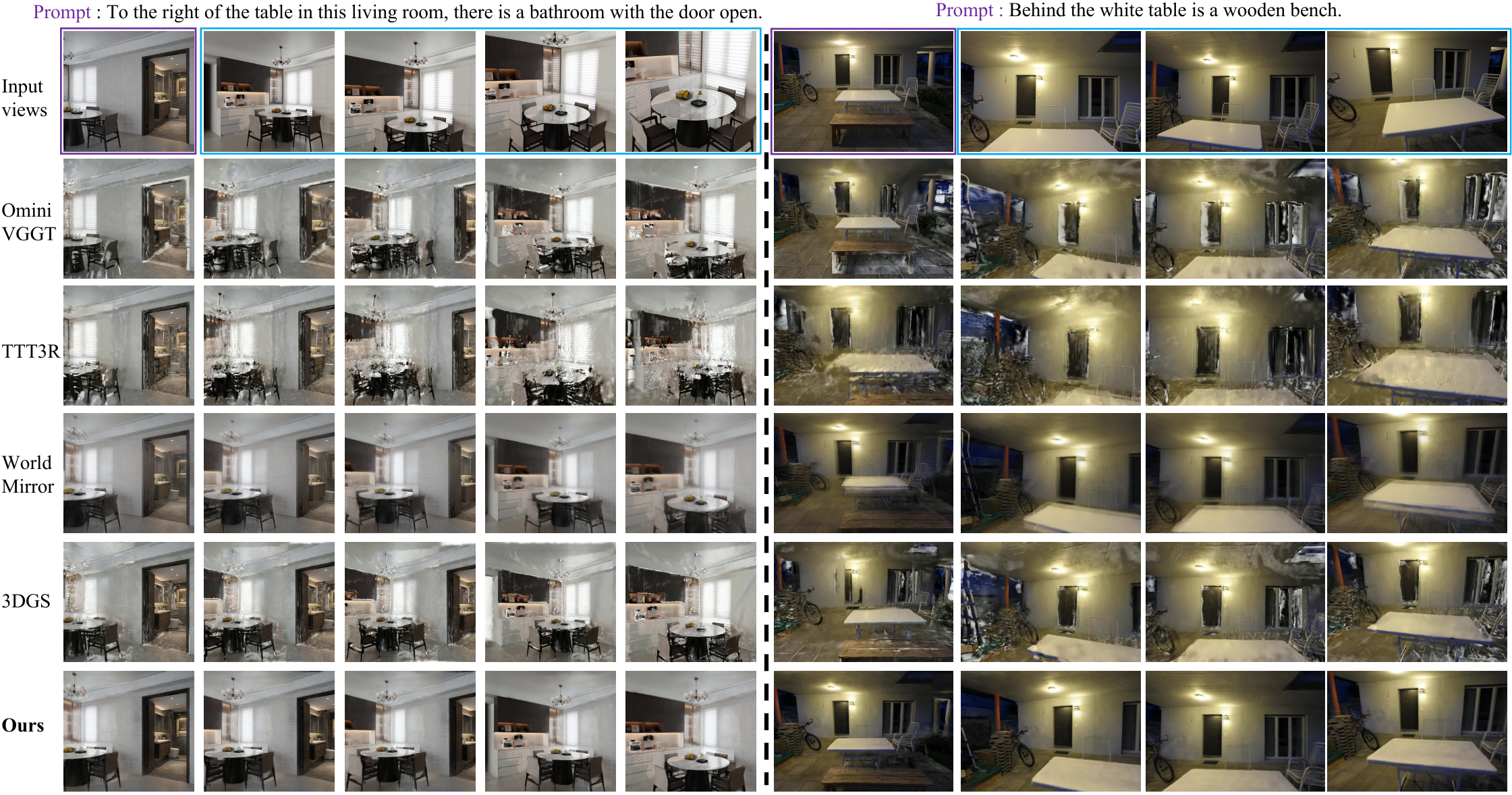}
  \caption{\textbf{Qualitative comparison} of scene expansion under misaligned insertion. We show an indoor scene from the Online collection (left) and an outdoor scene from the ETH dataset (right). The first row presents the generated inserted view (purple box) together with a subset of captured multi-view images from the original scene (blue box). Each subsequent row shows renderings at novel camera poses for different methods, illustrating both fidelity to the captured region and insertion satisfaction. Please refer to the \textit{media supplement} for video visual results.}
  \vspace{-15pt}
  \label{fig:qual_main}
\end{figure*}

\tit{Baselines}
\label{sec:baselines}
As no prior work explicitly studies 3D scene expansion via view insertion, we adapt representative reconstruction methods as baselines: WorldMirror~\cite{liu2025worldmirror}, OmniVGGT~\cite{peng2025omnivggt}, TTT3R~\cite{chen2025ttt3r}, and 3D Gaussian Splatting (3DGS)~\cite{kerbl20233d}. WorldMirror and OmniVGGT are feed-forward multimodal reconstruction models. For TTT3R, originally designed for dynamic reconstruction, we treat the inserted view as an additional time step. For 3DGS, we initialize the inserted-view pose with WorldMirror predictions and optimize the representation using image supervision.


\tit{Evaluation Protocol and Metrics} We evaluate two aspects: \emph{preservation} of the captured region and \emph{expansion} quality under view insertion.

\vspace{2pt}
\noindent\textit{Image-space fidelity metrics.}
To measure preservation in the captured region, we render the final reconstruction from the original camera poses and compare the renderings with the captured images using PSNR, SSIM~\cite{wang2004image}, and LPIPS~\cite{zhang2018unreasonable}.
PSNR and SSIM quantify pixel-level and structural similarity, while LPIPS measures perceptual similarity. 

\vspace{2pt}
\noindent\textit{VLM-based evaluation.}
Expansion quality is hard to quantify because the inserted view can be 3D-misaligned and the newly introduced region lacks ground truth. Following prior work on using vision-enabled GPT models as human-aligned evaluators~\cite{wu2024gpt}, we use a GPT model to score each result on a 5-point scale (1--5) for two criteria: GPT-IS (insertion satisfaction), evaluating whether the prompt-requested content is realized, and GPT-RQ (rendering quality), evaluating the visual quality and stability of navigation renderings, including potential degradation of the captured region.

\vspace{2pt}
\noindent\textit{Human evaluation.}
We conduct a user study with 39 participants, who rank all methods by insertion satisfaction (IS) and rendering quality (RQ). 
Ranks are converted to scores from 5 to 1, and H-IS/H-RQ report the averages over all participants and cases.


\subsection{Comparisons with State-of-the-Art Methods}
\label{sec:main_results}
\paragraph{Quantitative comparison.}
Tab.~\ref{tab:comparison} reports averaged results on the ETH dataset and the online collections. OmniVGGT and TTT3R perform poorly on both preservation and insertion metrics, indicating limited robustness once a misaligned inserted view is introduced. WorldMirror is more robust (19.30 PSNR, 0.614 SSIM) with strong human scores, but its GPT-RQ is modest (2.61), suggesting residual artifacts during navigation. The optimization-based 3DGS degrades severely when directly fitting the misaligned view. In contrast, SceneExpander achieves the best overall trade-off, improving preservation over WorldMirror (21.21 PSNR, 0.776 SSIM, 0.276 LPIPS) while increasing expansion quality (GPT-IS 3.72, GPT-RQ 3.05) and aligning with human judgments (H-IS 4.56, H-RQ 4.53).

\begin{table*}[!h]
  \centering
  \caption{\textbf{Quantitative comparison.} We report preservation on the captured region using image-space fidelity metrics (PSNR/SSIM/LPIPS) and expansion quality using LLM-based and human evaluation. GPT-IS/H-IS measure insertion satisfaction, and GPT-RQ/H-RQ measure overall rendering quality. $\uparrow$ higher is better, $\downarrow$ lower is better.} 
  \vspace{-5pt}
  \renewcommand{\arraystretch}{1.02}
  \begin{tabular*}{.98\linewidth}{l@{\extracolsep{\fill}}ccc|cc|cc}
    \hline
    Method & PSNR$\uparrow$ & SSIM$\uparrow$ & LPIPS$\downarrow$
    & GPT-IS$\uparrow$ & GPT-RQ$\uparrow$
    & H-IS$\uparrow$ & H-RQ$\uparrow$ \\
    \hline
    OminVGGT~\cite{peng2025omnivggt}      & 14.969 & 0.3647 & 0.370 & 3.16 & 2.66 & 2.26 & 2.40 \\
    TTT3R~\cite{chen2025ttt3r}             & 13.483 & 0.2747 & 0.457 & 2.94 & 2.38 & 1.72 & 1.71 \\
    WorldMirror~\cite{liu2025worldmirror} & 19.295     & 0.614     & 0.357     & 3.50 & 2.61 & 4.04 & 4.07 \\
    3DGS~\cite{kerbl20233d}               & 14.046 & 0.3045 & 0.418 & 3.16 & 2.61 & 2.40 & 2.32 \\
    \hline
    SceneExpander (Ours) & \textbf{21.206} & \textbf{0.776} & \textbf{0.276} & \textbf{3.72} & \textbf{3.05} & \textbf{4.56} & \textbf{4.53} \\
    \hline
  \end{tabular*}
  \label{tab:comparison}
  \vspace{-10pt}
\end{table*}

\paragraph{Qualitative comparison.}
Fig.~\ref{fig:qual_main} visualizes expanded reconstructions and navigation results. Although image generation models can understand the scene and the input prompt to produce an inserted view, the result is often only satisfactory at the pixel level, and it still exhibits noticeable misalignment in 3D consistency. OmniVGGT produces noisy artifacts and fails to reliably reconstruct the newly expanded region implied by the inserted view. Although TTT3R performs dynamic reconstruction, it still exhibits pronounced ghosting. WorldMirror also suffers from misalignment-induced artifacts, such as ghosting on the chair in the left example and on the table in the right example. While 3DGS can enforce global consistency through iterative optimization, forcing the model to fit a misaligned view inevitably introduces substantial noise and distortions. Compared with these baselines, our method better preserves the captured region and yields more plausible expanded geometry and appearance that explain the inserted view.

\subsection{Ablation Studies}
\label{sec:ablation}
We ablate key components of SceneExpander: (i) anchor distillation $\mathcal{L}_{\text{anchor}}$, (ii) subset sampling and geometry-perturbation augmentation, (iii) inserted-view self-distillation with insertion probability $p$, and (iv) stochastic restoration (Tab.~\ref{tab:ablation}). Adding anchor distillation substantially improves preservation on the captured region (higher PSNR/SSIM and lower LPIPS), confirming the importance of anchoring adaptation to captured-view geometric cues to avoid drifting. Subset sampling and augmentation further improve robustness and noticeably increase GPT-RQ, indicating more stable navigation renderings under misalignment. Introducing self-distillation improves insertion satisfaction and fidelity; using $p{=}0.5$ provides a better overall balance, while always inserting the generated view ($p{=}1.0$) achieves the highest GPT-IS but degrades preservation and GPT-RQ, suggesting overfitting to the inserted constraint. Finally, stochastic restoration yields the best preservation metrics while maintaining strong insertion satisfaction, resulting in the most favorable trade-off in the full model.

\begin{table}[!h]
  \centering
  \caption{\textbf{Ablation study.}
  Each row adds one component to SceneExpander.
  ``distill.'' denotes distillation, ``samp.'' denotes sampling,
  ``aug.'' denotes augmentation, and ``stoch. rest.'' denotes
  stochastic restoration.}
  \label{tab:ablation}

  \footnotesize
  \setlength{\tabcolsep}{1.1pt}
  \renewcommand{\arraystretch}{1.08}

  \begin{tabularx}{\columnwidth}{
    @{}
    >{\raggedright\arraybackslash}X
    |ccc|cc
    @{}
  }
    \hline
    Variant
    & PSNR
    & SSIM
    & LPIPS
    & \shortstack{GPT-IS}
    & \shortstack{GPT-RQ} \\
    \hline

    Baseline
    & 19.295 & 0.614 & 0.357 & 3.50 & 2.61 \\

    + Anchor distill.
    & 20.763 & 0.714 & 0.329 & 3.52 & 2.96 \\

    + Subset samp. \& aug.
    & 20.892 & 0.720 & 0.322 & 3.56 & \textbf{3.24} \\

    + Self-distill. ($p{=}0.5$)
    & 21.124 & 0.757 & 0.294 & 3.70 & 3.08 \\

    + Self-distill. ($p{=}1.0$)
    & 20.856 & 0.704 & 0.318 & \textbf{3.84} & 2.88 \\

    + Stoch. rest. (Full)
    & \textbf{21.206}
    & \textbf{0.776}
    & \textbf{0.276}
    & 3.72
    & 3.05 \\
    \hline
  \end{tabularx}

  \vspace{-15pt}
\end{table}

\begin{figure}[!h]
  \centering
  \includegraphics[width=\linewidth]{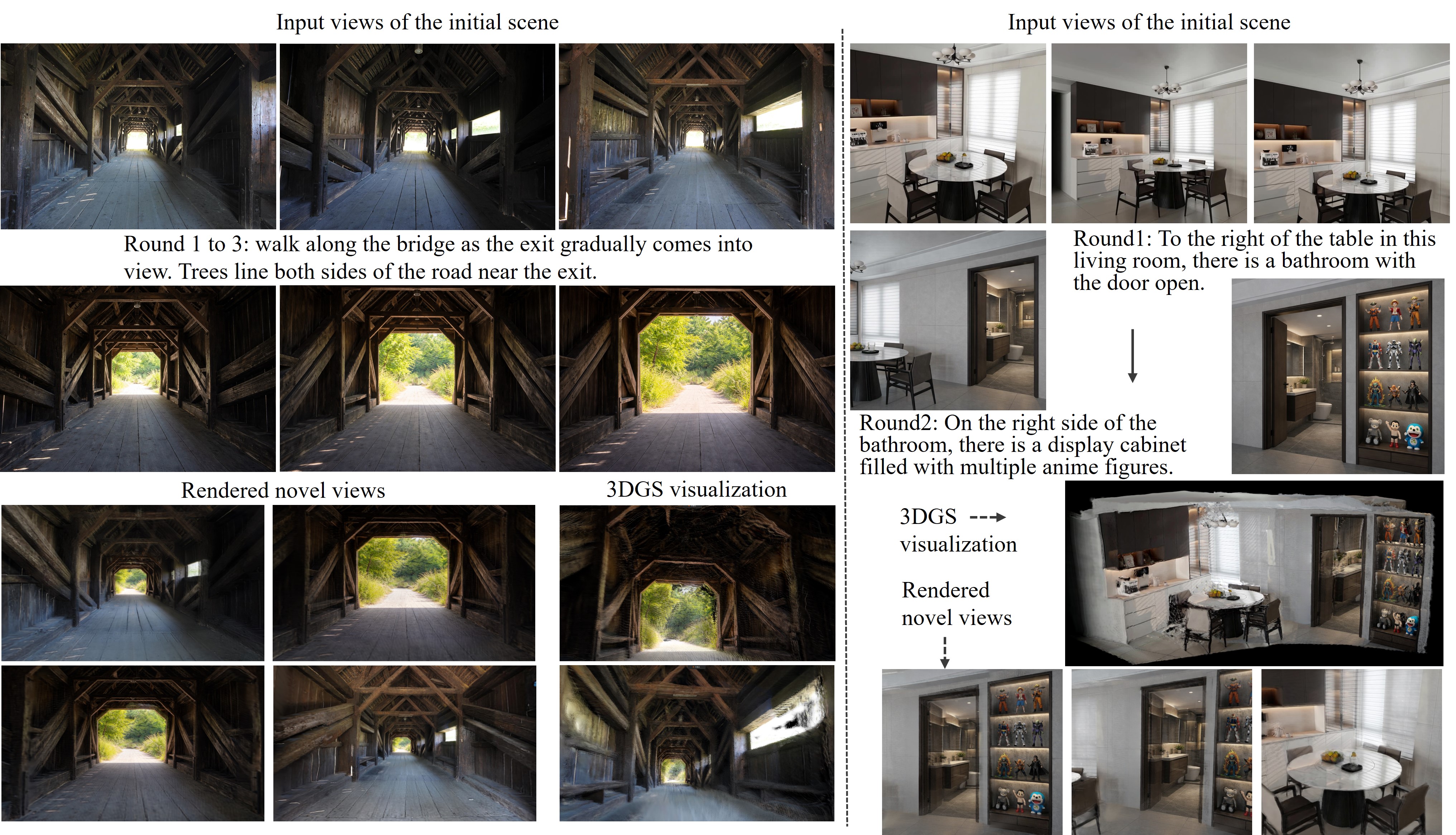}
    \caption{\textbf{Multi-round expansion.} 
    Starting from an initial captured scene, users can repeatedly insert new views to expand the reconstruction. 
    Across consecutive rounds, \ours\ progressively enlarges the 3D scene while preserving previously reconstructed regions.}
    \label{fig:multi_round}
    \vspace{-13pt}

\end{figure}

\subsection{Multi-Round Expansion}
\label{sec:multi_round_expansion}

We further evaluate \ours\ in an iterative world-building setting. 
At each round, a user specifies a new viewpoint or local content request, a new inserted view is synthesized, and \ours\ adapts the current reconstruction using the same test-time procedure as in the single-round setting. 
The adapted scene is then used as the starting point for the next round. As shown in Fig.~\ref{fig:multi_round}, consecutive insertions progressively expand the reconstructed scene and enable navigation into newly created regions. 
This suggests that single-view insertion can serve as a reusable local expansion operator for interactive 3D world building, allowing users to grow a scene step by step without regenerating the entire reconstruction from scratch.

\section{Conclusion}

We introduced \textbf{text-guided 3D scene expansion via free-form view insertion}, where a real captured scene is extended by a text-instantiated inserted view that may be geometrically misaligned with the original reconstruction. \ours\ robustly integrates such views through test-time adaptation with anchor distillation for captured-region preservation and inserted-view self-distillation for soft expansion. Experiments show improved preservation and expansion quality over representative baselines, while multi-round expansion results support iterative 3D world building. Future work can combine stronger 3D-consistent generation with robust integration to reduce error accumulation under severe conflicts.

\bibliography{main}

@article{loshchilov2017decoupled,
  title={Decoupled weight decay regularization},
  author={Loshchilov, Ilya and Hutter, Frank},
  journal={arXiv preprint arXiv:1711.05101},
  year={2017}
}

@article{wang2004image,
  title={Image quality assessment: from error visibility to structural similarity},
  author={Wang, Zhou and Bovik, Alan C and Sheikh, Hamid R and Simoncelli, Eero P},
  journal={IEEE transactions on image processing},
  volume={13},
  number={4},
  pages={600--612},
  year={2004},
  publisher={IEEE}
}

@inproceedings{zhang2018unreasonable,
  title={The unreasonable effectiveness of deep features as a perceptual metric},
  author={Zhang, Richard and Isola, Phillip and Efros, Alexei A and Shechtman, Eli and Wang, Oliver},
  booktitle={Proceedings of the IEEE conference on computer vision and pattern recognition},
  pages={586--595},
  year={2018}
}

@article{liu2025worldmirror,
  title={Worldmirror: Universal 3d world reconstruction with any-prior prompting},
  author={Liu, Yifan and Min, Zhiyuan and Wang, Zhenwei and Wu, Junta and Wang, Tengfei and Yuan, Yixuan and Luo, Yawei and Guo, Chunchao},
  journal={arXiv preprint arXiv:2510.10726},
  year={2025}
}

@inproceedings{wu2024gpt,
  title={Gpt-4v (ision) is a human-aligned evaluator for text-to-3d generation},
  author={Wu, Tong and Yang, Guandao and Li, Zhibing and Zhang, Kai and Liu, Ziwei and Guibas, Leonidas and Lin, Dahua and Wetzstein, Gordon},
  booktitle={Proceedings of the IEEE/CVF conference on computer vision and pattern recognition},
  pages={22227--22238},
  year={2024}
}

@inproceedings{schonberger2016structure,
  title={Structure-from-motion revisited},
  author={Schonberger, Johannes L and Frahm, Jan-Michael},
  booktitle={Proceedings of the IEEE conference on computer vision and pattern recognition},
  pages={4104--4113},
  year={2016}
}

@article{mildenhall2021nerf,
  title={Nerf: Representing scenes as neural radiance fields for view synthesis},
  author={Mildenhall, Ben and Srinivasan, Pratul P and Tancik, Matthew and Barron, Jonathan T and Ramamoorthi, Ravi and Ng, Ren},
  journal={Communications of the ACM},
  volume={65},
  number={1},
  pages={99--106},
  year={2021},
  publisher={ACM New York, NY, USA}
}

@article{kerbl20233d,
  title={3D Gaussian splatting for real-time radiance field rendering.},
  author={Kerbl, Bernhard and Kopanas, Georgios and Leimk{\"u}hler, Thomas and Drettakis, George},
  journal={ACM Trans. Graph.},
  volume={42},
  number={4},
  pages={139--1},
  year={2023}
}

@inproceedings{wang2025vggt,
  title={Vggt: Visual geometry grounded transformer},
  author={Wang, Jianyuan and Chen, Minghao and Karaev, Nikita and Vedaldi, Andrea and Rupprecht, Christian and Novotny, David},
  booktitle={Proceedings of the Computer Vision and Pattern Recognition Conference},
  pages={5294--5306},
  year={2025}
}

@article{peng2025omnivggt,
  title={OmniVGGT: Omni-Modality Driven Visual Geometry Grounded Transformer},
  author={Peng, Haosong and Li, Hao and Dai, Yalun and Lan, Yushi and Luo, Yihang and Qi, Tianyu and Zhang, Zhengshen and Zhan, Yufeng and Zhang, Junfei and Xu, Wenchao and others},
  journal={arXiv preprint arXiv:2511.10560},
  year={2025}
}

@article{chen2025ttt3r,
    title={TTT3R: 3D Reconstruction as Test-Time Training},
    author={Chen, Xingyu and Chen, Yue and Xiu, Yuliang and Geiger, Andreas and Chen, Anpei},
    journal={arXiv preprint arXiv:2509.26645},
    year={2025}
    }

@inproceedings{schneider2025worldexplorer,
  title={Worldexplorer: Towards generating fully navigable 3d scenes},
  author={Schneider, Manuel-Andreas and H{\"o}llein, Lukas and Nie{\ss}ner, Matthias},
  booktitle={Proceedings of the SIGGRAPH Asia 2025 Conference Papers},
  pages={1--11},
  year={2025}
}

@article{huang2025voyager,
  title={Voyager: Long-Range and World-Consistent Video Diffusion for Explorable 3D Scene Generation},
  author={Huang, Tianyu and Zheng, Wangguandong and Wang, Tengfei and Liu, Yuhao and Wang, Zhenwei and Wu, Junta and Jiang, Jie and Li, Hui and Lau, Rynson WH and Zuo, Wangmeng and Guo, Chunchao},
  journal={arXiv preprint arXiv:2506.04225},
  year={2025}
}

@article{mao2025yume,
  title={Yume-1.5: A Text-Controlled Interactive World Generation Model},
  author={Mao, Xiaofeng and Li, Zhen and Li, Chuanhao and Xu, Xiaojie and Ying, Kaining and He, Tong and Pang, Jiangmiao and Qiao, Yu and Zhang, Kaipeng},
  journal={arXiv preprint arXiv:2512.22096},
  year={2025}
}

@inproceedings{hollein2023text2room,
  title={Text2room: Extracting textured 3d meshes from 2d text-to-image models},
  author={H{\"o}llein, Lukas and Cao, Ang and Owens, Andrew and Johnson, Justin and Nie{\ss}ner, Matthias},
  booktitle={Proceedings of the IEEE/CVF International Conference on Computer Vision},
  pages={7909--7920},
  year={2023}
}

@article{zhang2024text2nerf,
  title={Text2nerf: Text-driven 3d scene generation with neural radiance fields},
  author={Zhang, Jingbo and Li, Xiaoyu and Wan, Ziyu and Wang, Can and Liao, Jing},
  journal={IEEE Transactions on Visualization and Computer Graphics},
  volume={30},
  number={12},
  pages={7749--7762},
  year={2024},
  publisher={IEEE}
}

@inproceedings{yang2024viewfusion,
  title={Viewfusion: Towards multi-view consistency via interpolated denoising},
  author={Yang, Xianghui and Zuo, Yan and Ramasinghe, Sameera and Bazzani, Loris and Avraham, Gil and van den Hengel, Anton},
  booktitle={Proceedings of the IEEE/CVF Conference on Computer Vision and Pattern Recognition},
  pages={9870--9880},
  year={2024}
}

@inproceedings{cao2025mvgenmaster,
  title={MVGenMaster: Scaling Multi-View Generation from Any Image via 3D Priors Enhanced Diffusion Model},
  author={Cao, Chenjie and Yu, Chaohui and Liu, Shang and Wang, Fan and Xue, Xiangyang and Fu, Yanwei},
  booktitle={Proceedings of the Computer Vision and Pattern Recognition Conference},
  pages={6045--6056},
  year={2025}
}

@inproceedings{haque2023instruct,
  title={Instruct-nerf2nerf: Editing 3d scenes with instructions},
  author={Haque, Ayaan and Tancik, Matthew and Efros, Alexei A and Holynski, Aleksander and Kanazawa, Angjoo},
  booktitle={Proceedings of the IEEE/CVF international conference on computer vision},
  pages={19740--19750},
  year={2023}
}

@misc{igs2gs,
 author = {Vachha, Cyrus and Haque, Ayaan},
 title = {Instruct-GS2GS: Editing 3D Gaussian Splats with Instructions},
 year = {2024},
 url = {https://instruct-gs2gs.github.io/}
}

@inproceedings{chen2024gaussianeditor,
  title={Gaussianeditor: Swift and controllable 3d editing with gaussian splatting},
  author={Chen, Yiwen and Chen, Zilong and Zhang, Chi and Wang, Feng and Yang, Xiaofeng and Wang, Yikai and Cai, Zhongang and Yang, Lei and Liu, Huaping and Lin, Guosheng},
  booktitle={Proceedings of the IEEE/CVF conference on computer vision and pattern recognition},
  pages={21476--21485},
  year={2024}
}

@inproceedings{wang2024view,
  title={View-consistent 3d editing with gaussian splatting},
  author={Wang, Yuxuan and Yi, Xuanyu and Wu, Zike and Zhao, Na and Chen, Long and Zhang, Hanwang},
  booktitle={European conference on computer vision},
  pages={404--420},
  year={2024},
  organization={Springer}
}

@inproceedings{lee2025editsplat,
  title={Editsplat: Multi-view fusion and attention-guided optimization for view-consistent 3d scene editing with 3d gaussian splatting},
  author={Lee, Dong In and Park, Hyeongcheol and Seo, Jiyoung and Park, Eunbyung and Park, Hyunje and Baek, Ha Dam and Shin, Sangheon and Kim, Sangmin and Kim, Sangpil},
  booktitle={Proceedings of the Computer Vision and Pattern Recognition Conference},
  pages={11135--11145},
  year={2025}
}

@article{chen2025fast,
  title={Fast Multi-view Consistent 3D Editing with Video Priors},
  author={Chen, Liyi and Li, Ruihuang and Zhang, Guowen and Wang, Pengfei and Zhang, Lei},
  journal={arXiv preprint arXiv:2511.23172},
  year={2025}
}

@article{dai2025fantasyworld,
  title={Fantasyworld: Geometry-consistent world modeling via unified video and 3d prediction},
  author={Dai, Yixiang and Jiang, Fan and Wang, Chiyu and Xu, Mu and Qi, Yonggang},
  journal={arXiv preprint arXiv:2509.21657},
  year={2025}
}

@inproceedings{schops2017multi,
  title={A multi-view stereo benchmark with high-resolution images and multi-camera videos},
  author={Schops, Thomas and Schonberger, Johannes L and Galliani, Silvano and Sattler, Torsten and Schindler, Konrad and Pollefeys, Marc and Geiger, Andreas},
  booktitle={Proceedings of the IEEE conference on computer vision and pattern recognition},
  pages={3260--3269},
  year={2017}
}

@inproceedings{zhuang2023dreameditor,
  title={Dreameditor: Text-driven 3d scene editing with neural fields},
  author={Zhuang, Jingyu and Wang, Chen and Lin, Liang and Liu, Lingjie and Li, Guanbin},
  booktitle={SIGGRAPH Asia 2023 conference papers},
  pages={1--10},
  year={2023}
}

@article{zhuang2024tip,
  title={Tip-editor: An accurate 3d editor following both text-prompts and image-prompts},
  author={Zhuang, Jingyu and Kang, Di and Cao, Yan-Pei and Li, Guanbin and Lin, Liang and Shan, Ying},
  journal={arXiv preprint arXiv:2401.14828},
  year={2024}
}

@article{wu2024gaussctrl,
  title={GaussCtrl: multi-view consistent text-driven 3D Gaussian splatting editing},
  author={Wu, Jing and Bian, Jia-Wang and Li, Xinghui and Wang, Guangrun and Reid, Ian and Torr, Philip and Prisacariu, Victor Adrian},
  journal={arXiv preprint arXiv:2403.08733},
  year={2024}
}

@inproceedings{he2025vton,
  title={Vton 360: High-fidelity virtual try-on from any viewing direction},
  author={He, Zijian and Ning, Yuwei and Qin, Yipeng and Wang, Guangrun and Yang, Sibei and Lin, Liang and Li, Guanbin},
  booktitle={Proceedings of the IEEE/CVF Conference on Computer Vision and Pattern Recognition},
  pages={26388--26398},
  year={2025}
}

@article{zheng2025splatpainter,
  title={SplatPainter: Interactive Authoring of 3D Gaussians from 2D Edits via Test-Time Training},
  author={Zheng, Yang and Tan, Hao and Zhang, Kai and Wang, Peng and Guibas, Leonidas and Wetzstein, Gordon and Yifan, Wang},
  journal={arXiv preprint arXiv:2512.05354},
  year={2025}
}

\newpage
\clearpage
\setcounter{page}{1}
\appendix

\section*{Appendix}

This appendix complements the main paper with additional qualitative results, implementation details, inserted-view generation and evaluation protocols, memory analysis, broader-impact discussion, and the full test-time adaptation algorithm of \ours.

\subsection{More Results and Limitation}
\label{sec:more_results}
\begin{figure*}[h]
  \centering
  \includegraphics[width=0.9\textwidth]{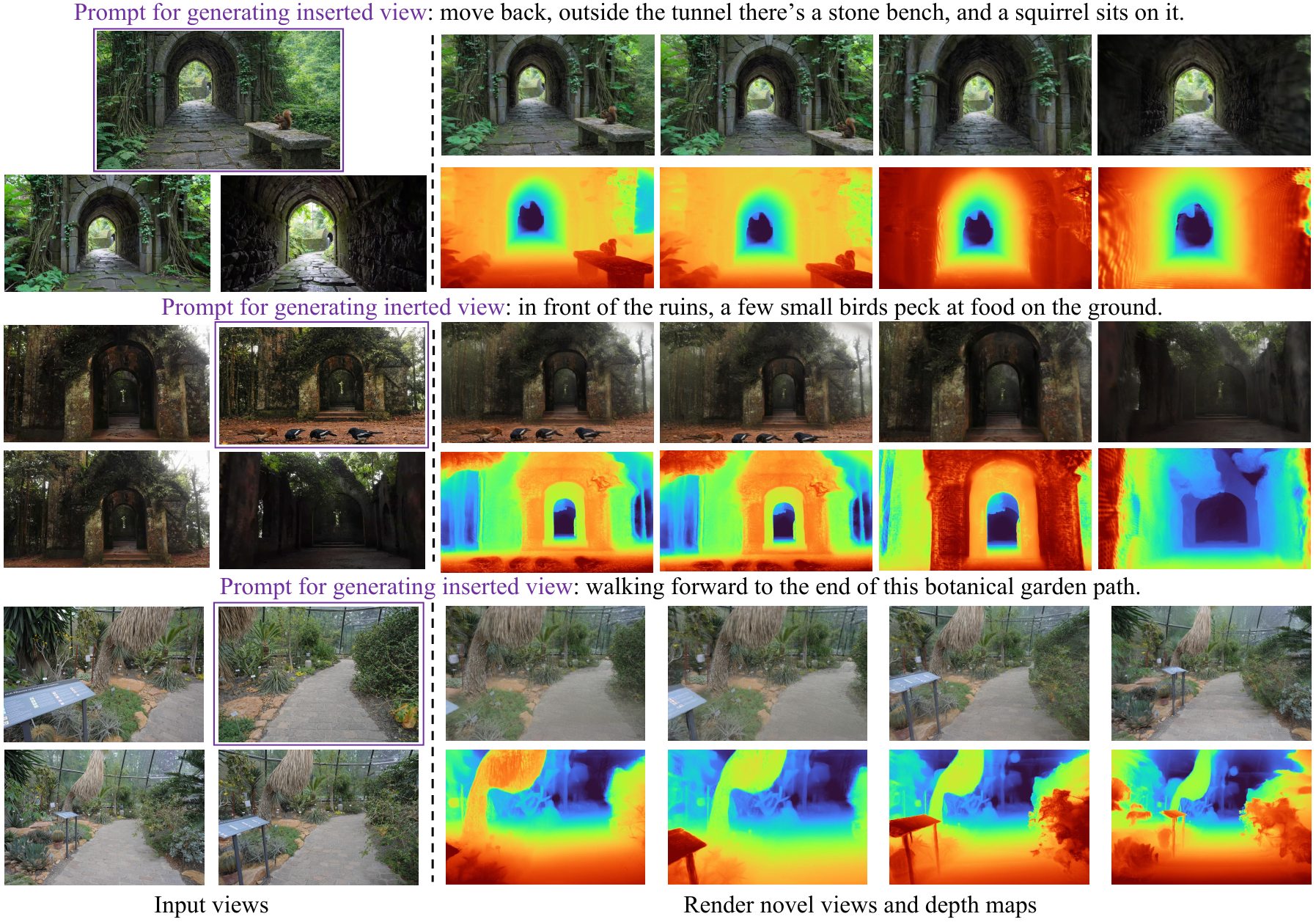}
  \caption{\textbf{More cases of SceneExpander.}
  Left of the dashed line: input images, where the purple dashed box
  indicates the generated inserted view and the remaining images are
  captured views. The second and third cases show a representative
  subset of captured views for brevity. Right of the dashed line:
  novel views and corresponding depth maps rendered by camera
  navigation in the expanded scene.}
  \label{fig:more}
\end{figure*}

We provide additional indoor and outdoor examples in Fig.~\ref{fig:more} to illustrate the motivation of interactive world building via view insertion. The first example mimics human exploration of a forest tunnel by extending the scene to include a squirrel on a stone bench outside the tunnel. The second expands an outdoor scene beyond the captured boundary to incorporate a bird pecking at food. The third continues a walkway in a botanical garden into an unseen region by extending the road structure. Renderings from camera navigation in the expanded reconstructions, together with the corresponding depth maps, show that SceneExpander can integrate the inserted view while maintaining coherent geometry and appearance for navigation.

\begin{figure*}[h]
    \centering
    \includegraphics[width=0.95\textwidth]{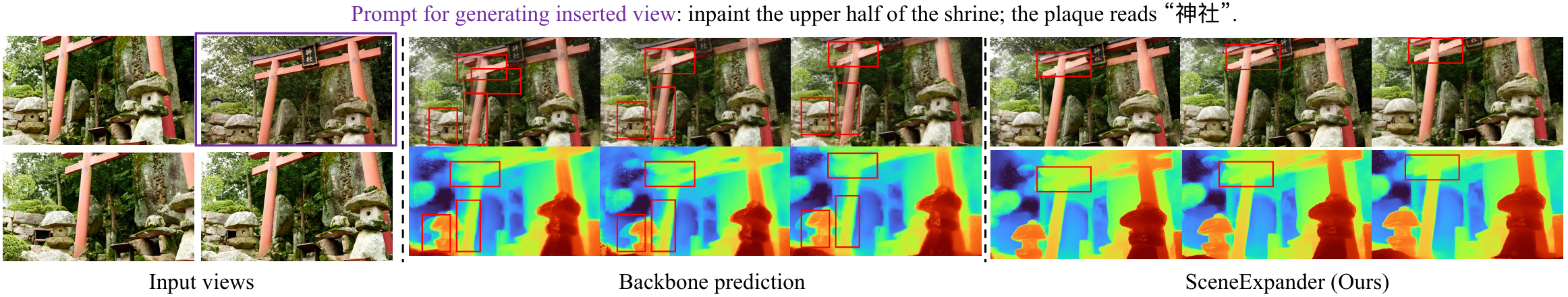}
    \caption{\textbf{A failure case of \ours.}
    The purple box shows the generated inserted view, and the red boxes highlight artifacts.
    \ours\ alleviates misalignment on the two side pillars and the left stone pedestal, but still produces errors on the top crossbeam.}
    \label{fig:failure}
\end{figure*}

Although SceneExpander mitigates geometric conflicts between the inserted view and the target scene through test-time adaptation, it may not fully resolve severe initial inconsistencies. As shown in Fig.~\ref{fig:failure}, although SceneExpander alleviates misalignment on the two side pillars and the left stone pedestal, noticeable errors remain on the plaque-supporting beam.


\subsection{Implementation Details}
\label{sec:impl}
We use the vision-language model GPT-5.2 to synthesize the inserted view in our expansion workflow, and to serve as the evaluator for our GPT-based metrics. We represent scenes using 3D Gaussian Splatting~\cite{kerbl20233d} with a differentiable renderer. During test-time adaptation, we insert $I_g$ with probability $p=0.5$. We optimize all parameters using AdamW~\cite{loshchilov2017decoupled} with a learning rate of $1\times10^{-5}$. We set $\alpha_D=\alpha_N=0.2$ for the anchor loss and $\beta_D=\beta_N=1.0$ for the self-distillation loss. The final objective uses $\lambda_A=1.0$, $\lambda_G=5\times10^{-2}$, and $\lambda_{\mathrm{reg}}=1\times10^{-4}$, and we set the stochastic restoration rate to $r=1\times10^{-3}$.

\subsection{Memory Consumption and Efficiency} \label{app:efficiency} We further report the computational cost of \ours\ during test-time adaptation. All measurements are conducted on a single NVIDIA A800 GPU using the same setting as the main experiments. For each scene, \ours\ performs 400 adaptation iterations. On average, the adaptation takes 12.4 minutes per scene and reaches 45.0GB peak GPU memory.

\subsection{Inserted View Generation Details}
\label{sec:inserted_view_generation}

We generate the inserted view on ETH dataset and online collections using a vision-language model conditioned on an exploration-style instruction. We use the following prompt template, where the bracketed fields are placeholders filled per case:
\begin{quote}\small
You are exploring. Please generate a new viewpoint following the requirements: [CAMERA\_MOTION], [NEW\_OBJECT].
\end{quote}
Here, [CAMERA\_MOTION] describes how the camera should move, \\
and [NEW\_OBJECT] specifies the new object or content that should appear in the inserted view. For example, the prompt of the first case in Fig.~\ref{fig:teaser} is:
\begin{quote}\small
You are exploring. Please generate a new viewpoint following the requirements: move back; outside the tunnel, there is a stone bench, and a squirrel sits on it.
\end{quote}
We generate multiple candidates and manually select the one that best matches the instruction \emph{visually} for insertion in our workflow.

\subsection{User Study Details}
\label{sec:user_study_details}

\begin{figure}[h]
  \centering
  \includegraphics[width=\linewidth]{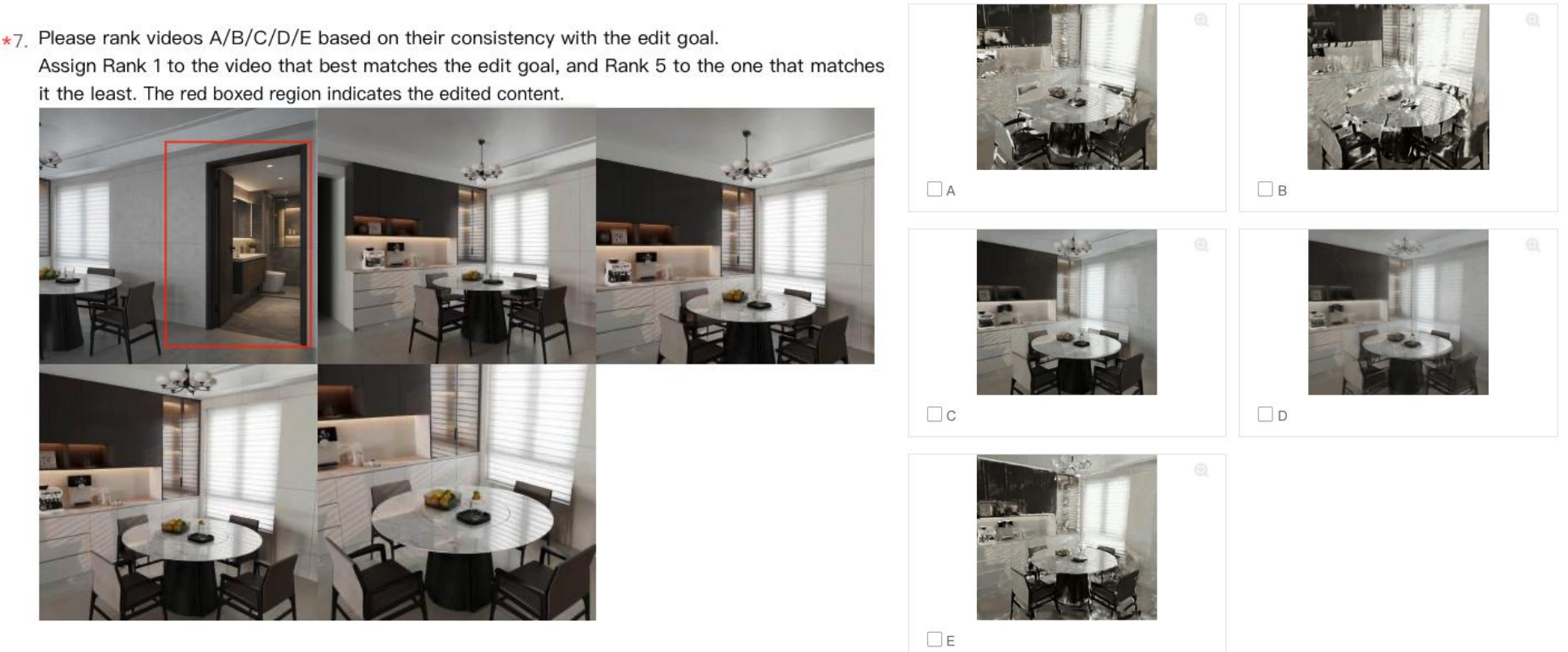}
  \caption{\textbf{User study interface.} Participants rank videos A–E by insertion satisfaction (H-IS) based on consistency with the edit goal, with the edited region highlighted in red. Options A–E include our method and four baselines, shown in a randomized order. The H-RQ interface follows the same format.}
  \label{fig:user_study_ui}
  \vspace{-8pt}
\end{figure}

We conduct a user study with 39 participants to evaluate insertion satisfaction (H-IS) and rendering quality (H-RQ). For each case, participants view the camera-navigation renderings from all methods and rank them from best to worst under each criterion. H-IS focuses on whether the prompt-requested content or new region is correctly realized and integrated, while H-RQ focuses on overall visual quality and stability, penalizing artifacts such as ghosting, noise, and distortions. We convert ranks to scores by assigning 5 to the top-ranked method and 1 to the bottom-ranked method, and report the average over all participants and cases. Fig.~\ref{fig:user_study_ui} shows the voting interface for H-IS, and the interface for H-RQ follows the same format.

\subsection{VLM-based Evaluation Details}
\label{sec:gpt_eval_details}

\begin{figure*}[h]
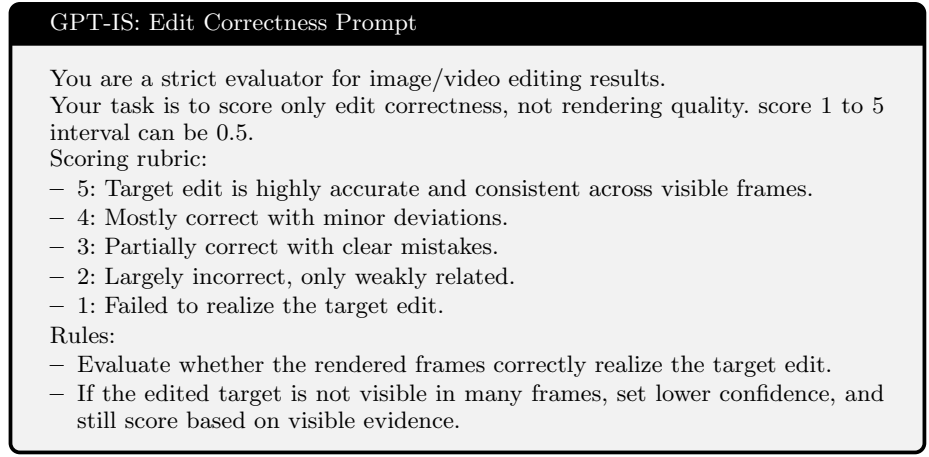

\begin{tcolorbox}[colback=black!5!white,colframe=black!75!black,title=GPT-IS: Edit Correctness Prompt]
\footnotesize
\setlength{\parskip}{0pt}
\setlength{\parindent}{0pt}
\setlength{\baselineskip}{0.92\baselineskip}
\setlist[itemize]{topsep=1pt,itemsep=0pt,parsep=0pt,partopsep=0pt,leftmargin=*}
You are a strict evaluator for image/video editing results.\\
Your task is to score only edit correctness, not rendering quality. score 1 to 5 interval can be 0.5.\\
Scoring rubric:
\begin{itemize}\setlength\itemsep{1pt}
\item 5: Target edit is highly accurate and consistent across visible frames.
\item 4: Mostly correct with minor deviations.
\item 3: Partially correct with clear mistakes.
\item 2: Largely incorrect, only weakly related.
\item 1: Failed to realize the target edit.
\end{itemize}
Rules:
\begin{itemize}\setlength\itemsep{1pt}
\item Evaluate whether the rendered frames correctly realize the target edit.
\item If the edited target is not visible in many frames, set lower confidence, and still score based on visible evidence.
\end{itemize}
\end{tcolorbox}
\caption{Prompt for VLM-based evaluation on insertion satisfaction.}
\label{fig:prompt}
\vspace{-10pt}
\end{figure*}

\begin{figure*}[h]
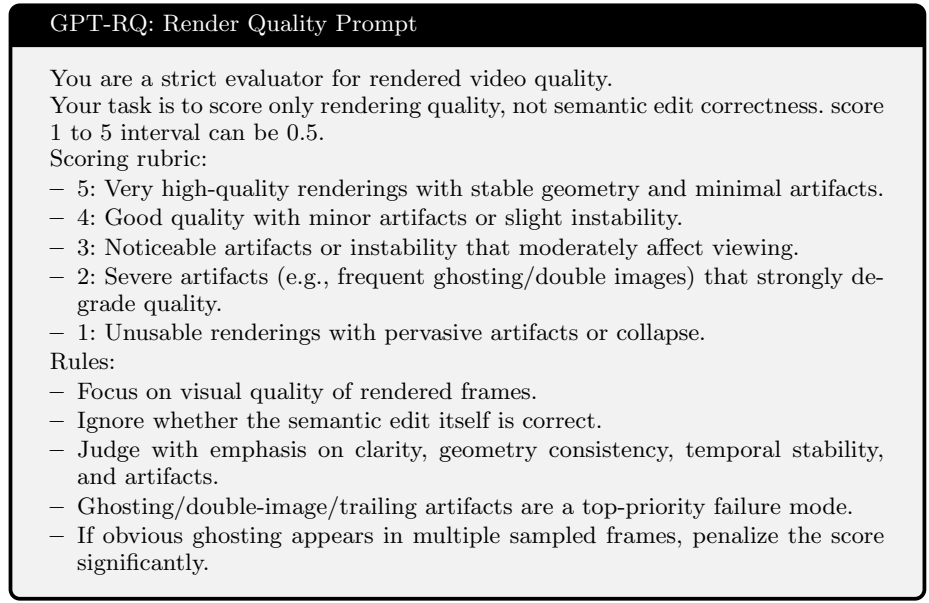

\begin{tcolorbox}[colback=black!5!white,colframe=black!75!black,title=GPT-RQ: Render Quality Prompt]
\footnotesize
\setlength{\parskip}{0pt}
\setlength{\parindent}{0pt}
\setlength{\baselineskip}{0.92\baselineskip}
\setlist[itemize]{topsep=1pt,itemsep=0pt,parsep=0pt,partopsep=0pt,leftmargin=*}
You are a strict evaluator for rendered video quality.\\
Your task is to score only rendering quality, not semantic edit correctness. score 1 to 5 interval can be 0.5.\\
Scoring rubric:
\begin{itemize}\setlength\itemsep{1pt}
\item 5: Very high-quality renderings with stable geometry and minimal artifacts.
\item 4: Good quality with minor artifacts or slight instability.
\item 3: Noticeable artifacts or instability that moderately affect viewing.
\item 2: Severe artifacts (e.g., frequent ghosting/double images) that strongly degrade quality.
\item 1: Unusable renderings with pervasive artifacts or collapse.
\end{itemize}
Rules:
\begin{itemize}\setlength\itemsep{1pt}
\item Focus on visual quality of rendered frames.
\item Ignore whether the semantic edit itself is correct.
\item Judge with emphasis on clarity, geometry consistency, temporal stability, and artifacts.
\item Ghosting/double-image/trailing artifacts are a top-priority failure mode.
\item If obvious ghosting appears in multiple sampled frames, penalize the score significantly.
\end{itemize}
\end{tcolorbox}
\caption{Prompt for VLM-based evaluation on rendering quality.}
\label{fig:prompt2}
\vspace{-10pt}
\end{figure*}

We use GPT-5.2 as a vision-language evaluator to assess results from the expanded 3D scene reconstruction. For each test case, we provide GPT with the target generated image and sampled rendered video frames produced by camera navigation in the reconstructed expanded scene. We compute two scores: insertion satisfaction (GPT-IS) and render quality (GPT-RQ), using the prompts shown in Fig.~\ref{fig:prompt} and Fig.~\ref{fig:prompt2}.

\subsection{Broader Impact}
SceneExpander supports a practical world-building workflow that expands a real captured scene by integrating a prompt-generated inserted view, which can reduce manual 3D authoring and benefit content creation, simulation, and interactive applications. However, expanded regions may contain plausible but incorrect geometry or hallucinated objects, and the workflow could be misused to fabricate 3D evidence by inserting non-existent content. We therefore recommend treating expanded areas as synthetic, clearly labeling generated insertions, and avoiding high-stakes use without additional verification.

\subsection{Future Work}
Future directions include long-horizon multi-step expansion with automatic view planning and quality control, uncertainty-aware constraint weighting to better handle severe misalignment, and incorporating stronger structural priors (e.g., semantics or layout) to improve coherence in newly expanded regions.

\subsection{Algorithm Details}
\label{app:algorithm}

Algorithm~\ref{alg:sceneexpander} summarizes the test-time adaptation procedure of \ours.
Captured views provide preservation constraints through anchor distillation, while the inserted view is sampled with probability $p$ and used as a soft expansion constraint through EMA self-distillation.

\clearpage
\onecolumn
\begin{algorithm*}[t]
\caption{\ours\ Test-Time Adaptation}
\label{alg:sceneexpander}
\small
\begin{algorithmic}[1]
\Require Captured views $\mathcal{C}=\{I_i\}_{i=1}^n$, inserted view $I_g$, steps $T$
\Require Insertion probability $p$, EMA momentum $\mu$, restore rate $r$, restore interval $K$
\State Initialize student $\theta \leftarrow \theta_0$
\State Initialize frozen anchor teacher $\theta_A \leftarrow \theta_0$
\State Initialize EMA teacher $\theta_G^{(0)} \leftarrow \theta_0$
\State Compute anchor references:
\Statex \hspace{\algorithmicindent}
$\{(\bar{\mathbf{P}}_i,\bar{D}_i,\bar{N}_i)\}_{i=1}^{n}
\leftarrow f_{\theta_A}(\mathcal{C})$
\For{$t=1$ to $T$}
    \State Sample captured indices $\mathcal{V}^{(t)} \subset \{1,\ldots,n\}$
    \State Form $\mathcal{B}^{(t)}=\{Aug(I_i)\}_{i\in\mathcal{V}^{(t)}}$
    \State Add $I_g$ to $\mathcal{B}^{(t)}$ with probability $p$
    \State Predict captured-view outputs:
    $\{(\widehat{\mathbf{P}}_i,\widehat{D}_i,\widehat{N}_i)\}_{i\in\mathcal{V}^{(t)}}
    \leftarrow f_{\theta}(\mathcal{B}^{(t)})|_{\mathcal{V}^{(t)}}$
    \State Compute $\mathcal{L}_{\mathrm{anchor}}$ using Eq.~\eqref{eq:anchor}
    \If{$I_g \in \mathcal{B}^{(t)}$}
        \State Predict student output on $I_g$:
        $(\widehat{\mathbf{P}}_g,\widehat{D}_g,\widehat{N}_g)
        \leftarrow f_{\theta}(\mathcal{B}^{(t)})|_g$
        \State Predict EMA target on $I_g$:
        $(\bar{\mathbf{P}}_g,\bar{D}_g,\bar{N}_g)
        \leftarrow f_{\theta_G^{(t-1)}}(\mathcal{B}^{(t)})|_g$
        \State Compute $\mathcal{L}_{\mathrm{gen}}$ using Eq.~\eqref{eq:gen}
    \Else
        \State $\mathcal{L}_{\mathrm{gen}} \leftarrow 0$
    \EndIf
    \State Update $\theta$ by minimizing:
    $\lambda_A\mathcal{L}_{\mathrm{anchor}}
    +\lambda_G\mathcal{L}_{\mathrm{gen}}
    +\lambda_{\mathrm{reg}}\mathcal{L}_{\mathrm{reg}}$
    \State Update EMA:
    $\theta_G^{(t)} \leftarrow \mu\theta_G^{(t-1)}+(1-\mu)\theta$
    \If{$t \bmod K = 0$}
        \State $\theta \leftarrow \textsc{Restore}(\theta,\theta_0;r)$
    \EndIf
\EndFor
\State \Return adapted model $\theta^\star \leftarrow \theta$
\end{algorithmic}
\label{alg:cotta_sceneexpander_compact}
\end{algorithm*}

\end{document}